\documentclass{article}

\usepackage{arxiv}
\usepackage[utf8]{inputenc}
\usepackage[T1]{fontenc}
\usepackage[hidelinks]{hyperref}
\usepackage{url}
\usepackage{booktabs}
\usepackage{amsfonts}
\usepackage{nicefrac}
\usepackage{microtype}
\usepackage{lipsum}
\usepackage{graphicx}
\usepackage[numbers, square]{natbib}
\usepackage{doi}
\usepackage[table]{xcolor}
\usepackage{graphicx}
\usepackage{amssymb}
\usepackage{eurosym}
\usepackage{float}
\usepackage{latexsym}
\usepackage{a4}
\usepackage{listings}
\usepackage{tikz}
\usetikzlibrary{fit,arrows,quotes,calc,automata,positioning,backgrounds,arrows.meta,bending,shapes,snakes,matrix}
\usepackage{adjustbox}
\usepackage{tablefootnote}
\usepackage{amsmath}
\usepackage{subcaption}
\usepackage{xcolor}
\usepackage{pifont}
\usepackage[automake,acronym]{glossaries-extra}
\usepackage{setspace}
\usepackage{xparse}

\title{Prompting LLMs with content plans to enhance \\ the summarization of scientific articles}

\author{ \href{https://orcid.org/0000-0002-7401-5198}{\includegraphics[scale=0.06]{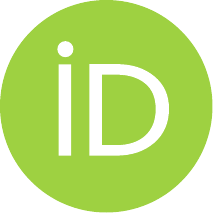}\hspace{1mm}Aldan~Creo} \\
	Singular Research Center on Intelligent Technologies (CiTIUS) \\
	University of Santiago de Compostela\\
	Santiago de Compostela, ES \\
	\texttt{aldan.creo@rai.usc.es} \\
	\And
	\href{https://orcid.org/0000-0001-7195-6155}{\includegraphics[scale=0.06]{orcid.pdf}\hspace{1mm}Manuel Lama} \\
	Singular Research Center on Intelligent Technologies (CiTIUS) \\
	University of Santiago de Compostela\\
	Santiago de Compostela, ES \\
	\texttt{manuel.lama@usc.es} \\
	\AND
	\href{https://orcid.org/0000-0002-8682-6772}{\includegraphics[scale=0.06]{orcid.pdf}\hspace{1mm}Juan C.~Vidal} \\
        Singular Research Center on Intelligent Technologies (CiTIUS) \\
	University of Santiago de Compostela \\
	Santiago de Compostela, ES \\
	\texttt{juan.vidal@usc.es} \\
}

\renewcommand{\headeright}{}
\renewcommand{\undertitle}{}
\renewcommand{\shorttitle}{Prompting LLMs with content plans to enhance the summarization of scientific articles}

\hypersetup{
pdftitle={Prompting LLMs with content plans to enhance the summarization of scientific articles},
pdfsubject={cs.AI, cs.CL},
pdfauthor={Aldan Creo, Manuel Lama, Juan C. Vidal},
pdfkeywords={Natural Language Processing, Artificial Intelligence, Summarization of Scientific Articles, Large Language Models},
}

\definecolor{ACMCMAGENTA}{HTML}{F600FF}
\definecolor{ACMCRED}{HTML}{FF0000}
\definecolor{ACMCBLUE}{HTML}{00AAEC} 
\newcommand\ACMCMAGENTA{ACMCMAGENTA}
\newcommand\ACMCRED{ACMCRED}
\newcommand\ACMCBLUE{ACMCBLUE}

\DeclareRobustCommand{\tecnica}[2][black]{\Gls{#2}}

\NewDocumentCommand{\tecnicaCombinada}{O{\ACMCMAGENTA} O{\ACMCBLUE} O{\ACMCRED} m m m}{
\tikz[baseline=(char.base)]{
\begin{scope}
\node[shape=rectangle, thick, inner sep=2.5pt, inner xsep=4pt, ] (charB) {\textcolor{white}{\footnotesize \bf \textsf{\hyperref[marker:#5]{#5}}}};
\node[shape=rectangle,node distance=0cm,left=of charB,inner sep=2.5pt, align=right, inner xsep=4pt, ] (charA) {\textcolor{white}{\footnotesize \bf \textsf{\hyperref[marker:#4]{#4}}}};
\node[shape=rectangle,,inner sep=2.5pt, inner xsep=4pt, node distance = 0cm, right=of charB, ] (charC) {\textcolor{white}{\footnotesize \bf \textsf{\hyperref[marker:#6]{#6}}}};
\begin{scope}[on background layer]
\node[inner sep=0pt, fit=(charA) (charB) (charC)] (container) {};
\fill[fill=#1, rounded corners=2pt] (charA.south east |- container.south) -- (container.south west) -- ($ (container.north west)!0.5!(container.south west) + (-0.2cm, 0.0cm)$) -- (container.north west) [sharp corners] -- (charA.east |- container.north) -- cycle;
\fill[fill=#2] (charA.east |- container.north) -- (charC.south west |- container.north) -- (charC.west |- container.south) -- (charA.south east |- container.south) -- cycle;
\fill[fill=#3, rounded corners=2pt] (charC.south west |- container.south) -- (container.south east) -- ($ (container.north east)!0.5!(container.south east) + (0.2cm, 0.0cm)$) -- (container.north east) [sharp corners] -- (charC.west |- container.north) -- cycle;
\draw[thick, rounded corners=2pt] (container.north west) -- (container.north east) -- ($ (container.north east)!0.5!(container.south east) + (0.2cm, 0.0cm)$) -- (container.south east) -- (container.south west) -- ($ (container.north west)!0.5!(container.south west) + (-0.2cm, 0.0cm)$) -- cycle;
\end{scope}
\end{scope}
}}

\makeglossaries

\newglossaryentry{BigBirdPegasus}
{
        name=BigBirdPegasus,
        description={\tecnica[\ACMCBLUE]{BigBirdPegasus} \cite{google/bigbird-pegasus-large-pubmed} combines Google's Pegasus \cite{Pegasus} model with the BigBird \cite{BigBird} sparse attention mechanism, allowing it to handle extremely long sequences. BigBird uses a combination of random, sliding window, and global attention to reduce the standard quadratic self-attention complexity. We incorporate \tecnica[\ACMCBLUE]{BigBirdPegasus} as it employs the most sophisticated attention scheme of models tested.}
}

\newglossaryentry{LT5-Base-Local}
{
        name=LT5-Base-Local,
        description={Variant of \Gls{LongT5} in base size employing sliding window attention only \cite{google/long-t5-local-base}.}
}

\newglossaryentry{LT5-Base-ETC}
{
        name=LT5-Base-ETC,
        description={Variant of \Gls{LongT5} in base size employing \Gls{ETC} attention \cite{google/long-t5-tglobal-base}.}
}

\newglossaryentry{LT5-Large-ETC}
{
        name=LT5-Large-ETC,
        description={Variant of \Gls{LongT5} in large size employing \Gls{ETC} attention \cite{google/long-t5-tglobal-large}.}
}

\newglossaryentry{LongT5}
{
        name=LongT5,
        description={LongT5 \cite{LongT5} is an extension of T5 \cite{T5} incorporating optimized sparse attention mechanisms to handle longer sequences beyond T5's 512 token limit. It allows input length up to 4096 tokens with either sliding window (\tecnica[\ACMCBLUE]{LT5-Base-Local}) or \Gls{ETC} attention (\tecnica[\ACMCBLUE]{LT5-Base-ETC} and \tecnica[\ACMCBLUE]{LT5-Large-ETC}).}
}

\newglossaryentry{LED}
{
        name=LED,
        description={A Longformer Encoder-Decoder is based on BART \cite{BART} while utilizing Longformer self-attention \cite{Longformer} to process long texts. Longformer combines sliding local attention windows with global attention from selected tokens. We integrate the base model to provide a comparison using an alternate attention approach from BigBird.}
}

\newglossaryentry{TF}
{
        name=TF,
        description={This unsupervised technique constructs prompts from the most frequently occurring non-stopwords in each article. TF provides a basic measure of term importance. Stopwords are filtered out using SpaCy \cite{spacy}.}
}

\newglossaryentry{TF-IDF}
{
        name=TF-IDF,
        description={A standard unsupervised statistic measuring how important words are to a document within a corpus. Words with high term frequency locally and low document frequency globally receive the highest scores. We extract the top \tecnica[\ACMCMAGENTA]{TF-IDF} terms for each section as prompts \cite{tf-idf}.}
}

\newglossaryentry{MeSH}
{
        name=MeSH,
        description={In biomedical scientific articles, manuscripts are indexed with MeSH (Medical Subject Headings) terms indicating key topics \cite{mesh}. They provide a comprehensive controlled vocabulary for medical concepts.}
}

\newglossaryentry{KeyBERT}
{
        name=KeyBERT,
        description={An unsupervised keyword extraction model based on distilled BERT \cite{devlin-etal-2019-bert} embeddings. We generate keywords from full input texts to form prompts, letting the model determine important terms rather than authors \cite{KeyBERT}.}
}

\newglossaryentry{Keywords}
{
        name=Keywords,
        description={The technique constructs prompts from author-provided keywords for each article. Author keywords offer an intuitive indicator of salient topical content. Prompts contain keywords separated by a delimiter.}
}

\newglossaryentry{I+D}
{
        name=I+D,
        description={Concatenation of the introduction and discussion texts, employed as a method to approximate the full text without incorporating the complexity of processing it entirely. This approach is motivated by the findings of \cite{bigpatent}, which demonstrates that the performance of summarization models utilizing both the introduction and discussion is comparable to those using the complete text. Given the cohesive overview expressed in the integration of introduction and discussion, our hypothesis is that prompts may yield lesser gains.}
}

\newglossaryentry{S-n/a}
{
        name=S-n/a,
        description={Concatenation of the introduction and discussion texts. When summarizing sections in isolation, we hypothesize that prompts may be more useful for recovering lost global context no longer expressed within single sections. We expect prompts may help more when summarizing sections independently.}
}

\newglossaryentry{S-w/a}
{
        name=S-w/a,
        description={Same as \tecnica[\ACMCRED]{S-n/a}, with the addition that the \underline{\textbf{i}}dentifier of the section type is prepended to the prompt, our hypothesis being that the additional context given may entail an increase in performance.}
}

\newglossaryentry{IMRAD}
{
        name=IMRAD,
        description={A commonly followed scheme for the structure of the abstracts of scientific articles, most prominently used in medical literature, featuring separate texts for the Introduction, Methods, Results and Discussion sections \cite{imrad-structure}.}
}

\newglossaryentry{ETC}
{
        name=ETC,
        description={Extended Transformer Construction (ETC) is an attention mechanism that utilizes both local attention and attention towards a set of global tokens \cite{ETC}.}
}

\begin{document}
\maketitle

\begin{abstract}
This paper presents novel prompting techniques to improve the performance of automatic summarization systems for scientific articles. Scientific article summarization is highly challenging due to the length and complexity of these documents. We conceive, implement, and evaluate prompting techniques that provide additional contextual information to guide summarization systems. Specifically, we feed summarizers with lists of key terms extracted from articles, such as author keywords or automatically generated keywords. Our techniques are tested with various summarization models and input texts. Results show performance gains, especially for smaller models summarizing sections separately. This evidences that prompting is a promising approach to overcoming the limitations of less powerful systems. Our findings introduce a new research direction of using prompts to aid smaller models.
\end{abstract}

% keywords can be removed

\section{Introduction}

Automatic text summarization is an active area of research dedicated to producing shortened versions of documents while retaining the most relevant information. Initial approaches to automatic summarization heavily leaned on extractive methods; however, most current state-of-the-art systems are based on abstractive summarization models, such as transformer architectures \cite{Vaswani2017}, which have demonstrated state-of-the-art results. These models are commonly implemented as encoder-decoder architectures, where the encoder builds representations of the input text, and the decoder generates the target summary. Extensive pretraining on large datasets equips these models with strong language modeling capabilities to generate fluent abstractive summaries.

While automatic summarization has applications across many domains involving large volumes of text, one notably challenging genre is scientific articles. Summarizing scientific articles poses difficulties beyond summarizing other document types due to these papers' great length and linguistic complexity. Scientific articles exhibit high variability in length. The highly technical vocabulary and complex discourse structures make summarization of scientific papers challenging even for state-of-the-art natural language processing systems \cite{Cohan2018, teufel-moens}.

Furthermore, scientific articles follow irregular organizational structures, unlike genres such as news with predictable templated content. In the biomedical domain focused on in this paper, a commonly followed scheme is the IMRAD structure (Introduction, Methods, Results, and Discussion) \cite{imrad-structure}. However, specifics may vary between subfields and journals, with sections further divided or additional sections present. This variability in structure poses difficulties for summarization systems to adapt. Consequently, abstracting scientific articles is acknowledged as a remarkably challenging domain within the field of automatic text summarization \cite{reviewAutoSumm}.

We introduce and investigate novel prompting techniques to improve the performance of state-of-the-art scientific article summarizers based on transformer architectures. In this paper, `prompting' refers to feeding summarizers with lists of key terms extracted from the input articles. Intuitively, supplying key terms may help focus summarizers on salient concepts to include in outputs. Our central hypothesis is that decoder prompting techniques will lead to gains in standard automatic evaluation metrics of summarization quality.

We conduct experiments applying our proposed prompting techniques to various state-of-the-art transformer-based summarization models. Our techniques are conceived to be easily obtainable for any input text, not relying on future knowledge as in some prior work \cite{EntityPrompts}. Specifically, we test five models covering a diversity of architectures: \Gls{LongT5} small and large variants employing different attention mechanisms, \tecnica[\ACMCBLUE]{LED} \cite{allenai/led-base-16384}, which uses Longformer attention \cite{Longformer}, and \tecnica[\ACMCBLUE]{BigBirdPegasus} \cite{google/bigbird-pegasus-large-pubmed} combining BigBird \cite{BigBird} and Pegasus \cite{Pegasus} architectures. We also experiment with different text inputs, either the concatenation of introduction and discussion text or sections separately, to determine whether prompts provide greater gains when summarizing sections in isolation.

Our results demonstrate consistent performance improvements from prompting techniques on smaller models, especially when summarizing sections independently. These models obtain ROUGE-1 score increases around $0.1$-$0.4$ when summarizing sections aided by prompts. In confusion testing, smaller models exhibit performance degradation when fed unrelated prompts, indicating reliance on prompt information. Taken together, these findings reveal that prompting is an effective approach to overcoming the fundamental limitations of smaller, less capable summarization systems. Rather than large models, lightweight models supplemented with prompts may be preferable in resource-constrained contexts like mobile devices.

Overall, the core contributions of this work are:

\begin{itemize}
\item We propose novel prompting techniques to provide key term context and enhance scientific literature summarizers.
\item Our techniques obtain prompts through simple extraction methods not requiring future knowledge, in contrast to prior work \cite{EntityPrompts}.
\item We perform extensive experiments on combinations of various state-of-the-art models and input text types, analyzing factors impacting prompt effectiveness.
\item Our results show enhanced performance when employing prompts for smaller summarization models, particularly in the context of section-level summarization. This underscores the potential of prompting as a promising technique to assist smaller models, particularly when computational resources are constrained.
\end{itemize}

The remainder of this paper is structured as follows. Section \ref{sec:related-work} reviews related prior work on which we build. Section \ref{sec:methods} presents our proposed methods for conceiving and evaluating prompting techniques. Section \ref{sec:results} details our experimental setup and results. Section \ref{sec:discussion} provides a discussion analyzing our findings. Section \ref{sec:future-work} proposes opportunities for future research. Finally, Section \ref{sec:conclusion} provides concluding remarks.

\section{Related Work}
\label{sec:related-work}

Automatically summarizing scientific articles is a long-studied challenge within automatic text summarization. Earlier conventional approaches relied heavily on extractive methods, where summarization involves identifying and extracting salient snippets, typically sentences, from the original document. For instance, one common family of techniques scored sentences based on statistical metrics like word frequencies to determine the importance of extraction \cite{yang2016, enfoque-extractivo-saggion, agrawal2019}. There have also been attempts to produce scientific article summaries by analyzing their citation contexts \cite{quazvinian-citation, cohan-citations, ronzano-citations}.

However, the current dominant paradigm has shifted decisively toward abstractive methods using neural network architectures. Our work focuses on techniques to enhance state-of-the-art abstractive scientific summarizers based on transformer models, which have become ubiquitous in natural language generation applications. Before discussing our proposed contributions, we first contextualize our work by summarizing prior studies on which we seek to build.

\subsection{Planning with Learned Entity Prompts}
\label{sed:related-work:prompting-with-named-entities}

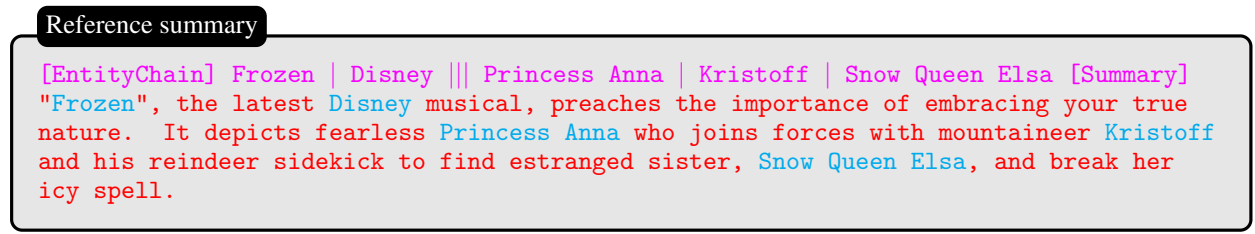
\begin{figure}[htbp]
    \centering
    \tikzstyle{mybox} = [draw=black, fill=gray!20, very thick,
        rectangle, rounded corners, align=center, inner sep=10pt, inner ysep=10pt]
    \tikzstyle{fancytitle} =[fill=black, text=white]
    \begin{tikzpicture}
    \node [mybox] (box){%
        \begin{minipage}{0.95\textwidth}%\centering
        \begin{ttfamily}\textcolor{\ACMCMAGENTA}{[EntityChain] Frozen $\vert$ Disney $\vert \vert \vert$ Princess Anna $\vert$ Kristoff $\vert$ Snow Queen Elsa [Summary]} \textcolor{\ACMCRED}{"\textcolor{\ACMCBLUE}{Frozen}", the latest \textcolor{\ACMCBLUE}{Disney} musical, preaches the importance of embracing your true nature. It depicts fearless \textcolor{\ACMCBLUE}{Princess Anna} who joins forces with mountaineer \textcolor{\ACMCBLUE}{Kristoff} and his reindeer sidekick to find estranged sister, \textcolor{\ACMCBLUE}{Snow Queen Elsa}, and break her icy spell.}
\end{ttfamily}
        \end{minipage}
    };
    \node[fancytitle, right=10pt, yshift=3pt, rounded corners] at (box.north west) {Reference summary};
    \end{tikzpicture}
\caption{Prompt example from \cite{EntityPrompts}. Prompt in \textcolor{\ACMCMAGENTA}{magenta}, summary in \textcolor{\ACMCRED}{red}, entities in \textcolor{\ACMCBLUE}{blue}.}
\label{fig:prompt-example}
\end{figure}

An especially relevant prior technique our work aims to build on is planning with learned entity prompts, proposed by \cite{EntityPrompts}. Their method trains scientific summarizers by providing an instruction composed of the named entities present in the reference summary, ordered by appearance. For example, as illustrated in Figure \ref{fig:prompt-example}, entities like `Frozen', `Princess Anna', or `Snow Queen Elsa' are fed to the decoder to control summary content.

Their training procedure involves first generating the prompt by extracting entities from the reference summary. The prompt is concatenated to the article text as the decoder input. At inference time, the model takes only the article as input and must generate both prompt and summary from scratch. According to their experiments, this approach improves faithfulness to included entities and reduces hallucination.

A clear limitation, however, is that accurately predicting key entities in advance to generate high-quality prompts is highly challenging, especially for complex scientific documents. The authors address this by further pretraining their model to generate entity chains from abstracts but do so only using scientific news summaries, which are far more straightforward than scientific articles. Directly producing entity prompts for long technical papers thus remains an unsolved difficulty.

Our work explores alternative prompting techniques that do not rely on advanced knowledge of key entities; instead, we extract prompts directly from input texts using simple, unsupervised methods. Models can thereby leverage prompts without strictly following specified entities. Our techniques are designed to be easily obtainable without needing additional complex models for prompt generation.

\subsection{Faceted Summarization}
\label{sec:related-work:faceted-summarization}

Another relevant line of work that inspired our methods is scientific summarization by individual paper sections rather than full text. Two prominent papers in this area are \cite{FacetSum} and \cite{structured-abstract-summarization}. The fundamental concept involves partitioning scientific articles into distinct sections and subsequently generating independent summaries for each section. These individual summaries are then amalgamated to form a comprehensive summary encapsulating the entire text. The key advantage is simplifying the summarization task by restricting focus to one section, which reduces complexity. Models can concentrate on concepts specific to individual sections rather than needing to capture the entire document.

In \cite{FacetSum}, the decoder input includes an instruction denoting which section type is being summarized. They train a single model on all sections while providing the section identifier. In contrast, \cite{structured-abstract-summarization} trains separate models for each section type. We similarly study section-level summarization but adopt the former approach of training singular models, as in our preliminary experiments, we did not observe clear benefits from specialized models.

A limitation acknowledged by both papers is that summarizing sections in isolation loses important global context, as sections relate closely to each other. Our prompting techniques aimed at providing contextual information may be particularly beneficial for section-focused summarization, which we analyze in our experiments.

Overall, prior work motivates the investigation of scientific article summarization using individual sections. Nevertheless, the absence of access to the complete text may result in the loss of contextual information that is not inherently evident in isolated sections. This information deficit may hinder the generation of a comprehensive summary. Our prompts containing key terms from the entire article could assist in recovering lost global context when summarizing sections individually.

\section{Methods}
\label{sec:methods}

This section details our proposed methods, which conceive and evaluate a set of novel prompting techniques to enhance the performance of state-of-the-art transformer models on scientific summarization across various settings. We describe our three key evaluation dimensions below.

\subsection{Prompting Technique Dimension}
\label{sec:methods:prompting-technique-dimension}

Our core dimension involves comparing various approaches for generating prompts to provide scientific summarizers with useful contextual information. As highlighted in Section \ref{sec:related-work}, a prior method by \cite{EntityPrompts} instructed models with chains of entities extracted from reference summaries. However, accurately predicting key entities in advance is highly challenging for complex documents like scientific articles.

Instead, our techniques provide lists of salient terms obtained through unsupervised extraction from input texts. The key hypothesis is that supplying relevant terms will help focus summarizers on important concepts to include in generated outputs. Crucially, our prompts are conceived to be easily obtainable without needing to involve additional models or knowledge of the structure of the summary to be generated.

In total, we conceive and evaluate five distinct prompting techniques:
\begin{itemize}
    \item \tecnica[\ACMCMAGENTA]{Keywords}, which consists of author-curated terms, not necessarily covering all topics.
    \item \tecnica[\ACMCMAGENTA]{MeSH}, providing a taxonomy of terms related to the article.
    \item \tecnica[\ACMCMAGENTA]{KeyBERT}, leveraging pretrained language representations for identifying salient terms.
    \item \tecnica[\ACMCMAGENTA]{TF}, taking the most frequent terms in the whole text.
    \item \tecnica[\ACMCMAGENTA]{TF-IDF}, selecting the terms that have a higher frequency with respect to the other sections in the text.
\end{itemize}

All our proposed techniques structure prompts similarly, starting with a delimiter token indicating prompt start, e.g., \texttt{[CONTENT]}, followed by the extracted key terms for that article section, ending with a \texttt{[SUMMARY]} token to denote summary start: \texttt{[CONTENT] term1 $|$ term2 $|$ ... termN [SUMMARY]}.

The core hypothesis is that training summarization models with these informative term prompts will allow scientific summarizers to better identify and focus on salient concepts to include in generated outputs. Models can learn to leverage supplied terms as useful guidance without having to rigidly follow exactly the provided keywords, which may not cover all necessary concepts. During training, reference summaries are concatenated after prompts to optimize models as usual through teacher forcing, while only prompts are given at inference time.

In summary, our experiments aim to determine whether these easily obtained general prompts listing salient terms can enhance summarizers' focus on core concepts to improve output quality without requiring strict instruction following or complex generative models for prompt production. We evaluate their impact when integrated with various state-of-the-art summarizers. By comparing a diversity of extraction techniques to generate prompts, we aim to determine what styles of key terms best provide useful context. We expect that effectiveness may vary with respect to the technique used.

\subsection{Model Dimension}
\label{sec:methods:model-dimension}

The second dimension we investigate involves studying our proposed prompting techniques integrated with a range of current state-of-the-art transformer models for scientific summarization. As highlighted in Section \ref{sec:related-work}, transformers have become dominant for abstractive summarization. We select a diversity of architectures to analyze prompting effects more robustly.

The core set of models we evaluate are:

\begin{itemize}
\item \tecnica[\ACMCBLUE]{LT5-Base-Local} \cite{google/long-t5-local-base}, \tecnica[\ACMCBLUE]{LT5-Base-ETC} \cite{google/long-t5-tglobal-base} and \tecnica[\ACMCBLUE]{LT5-Large-ETC} \cite{google/long-t5-tglobal-large}, three variants in size and attention architecture of the \Gls{LongT5} \cite{LongT5} model.
\item\tecnica[\ACMCBLUE]{LED}, Longformer Encoder-Decoder, combining the architecture of BART \cite{BART} with the Longformer \cite{Longformer} attention model.
\item \tecnica[\ACMCBLUE]{BigBirdPegasus}, applying the BigBird \cite{BigBird} sparse attention mechanism to Google's Pegasus \cite{Pegasus} model.
\end{itemize}

These choices cover a range of attention mechanisms, model sizes, and base architectures. \Gls{LongT5} allows comparing local against \Gls{ETC} attention, and base against large model size scaling; \tecnica[\ACMCBLUE]{LED} provides an alternate architecture through Longformer attention; and \tecnica[\ACMCBLUE]{BigBirdPegasus} employs the most complex attention. All models are fine-tuned on our scientific summarization dataset with and without prompts.

By integrating prompting techniques with multiple state-of-the-art summarizers, we aim to determine whether effectiveness generalizes across models or varies by architecture. We hypothesize that smaller models may benefit more from prompts providing helpful context compared to large models with greater representation capacity. Similarly, global attention may use prompts more effectively than models using sparse local attention only. Analyzing interactions between prompts and models is a key goal.

\subsection{Input Text Dimension}
\label{sec:methods:input-text-dimension}

The third dimension explored concerns the text inputs fed to the encoders. We study three main conditions:

\begin{itemize}
    \item \tecnica[\ACMCRED]{I+D}, where we take the concatenation of the \underline{\textbf{i}}ntroduction and \underline{\textbf{d}}iscussion sections.
    \item \tecnica[\ACMCRED]{S-n/a}, taking the texts of the different sections, summarising them separately, and later integrating the different summaries into a general summary, as described in Section \ref{sec:related-work:faceted-summarization}. The prompt for the summarization of each \underline{\textbf{s}}ection is \underline{\textbf{n}}ot \underline{\textbf{a}}nnotated with the type of the section.
    \item \tecnica[\ACMCRED]{S-w/a}, adopting the same approach as \tecnica[\ACMCRED]{S-n/a}, but prepending the prompt corresponding to each \underline{\textbf{s}}ection \underline{\textbf{w}}ith the \underline{\textbf{a}}nnotation of the type of the section (e.g., `Introduction', `Methods', \dots).
\end{itemize}

We hypothesize that in the case of \tecnica[\ACMCRED]{I+D}, prompts may provide lesser gains since this text already expresses a cohesive overview. On the other hand, when summarizing sections in isolation (\tecnica[\ACMCRED]{S-n/a} and \tecnica[\ACMCRED]{S-w/a}), prompts may be more helpful in recovering lost global context no longer expressed within single sections. We expect prompts may help more when summarizing sections independently.

Additionally, when generating section-level summaries, we compare either providing just the target section text alone (\tecnica[\ACMCRED]{S-n/a}), or prepending the prompt with an explicit token denoting the section type, e.g., \texttt{[INTRODUCTION]} (\tecnica[\ACMCRED]{S-w/a}). We hypothesize that explicitly indicating the section type could further improve results by focusing models on that section's expected content.

By evaluating the three settings, we aim to understand when contextual information from prompts provides the most significant gains. We expect faceted summarization may benefit most from supplementary global context through prompts.

\section{Results}
\label{sec:results}

This section describes our experiments evaluating the conceived prompting techniques integrated with various state-of-the-art summarization models and input texts. We first detail the dataset used, preprocessing performed, training methodology, and evaluation protocol. We then present the results.

\subsection{Datasets}

To train and evaluate scientific summarization systems, we use a dataset of open-access biomedical papers from PubMed Central \cite{pubmed-baseline}. We focus on the biomedical domain due to the large volume of openly available papers with structured abstracts summarizing key sections.

We extract a subset of 11,614 articles satisfying filtering criteria designed to obtain high-quality training examples from the complete PubMed Central corpus. For inclusion, papers must contain full text, author keywords, and structured abstracts with identifiable Introduction, Methods, Results, and Discussion (\Gls{IMRAD}) sections modeling standard scientific manuscript organization \cite{imrad-structure}. Articles deviating over two standard deviations from the mean length are excluded as outliers. This exclusion ensures that adequately sized input texts remain available for training summarization models.

In the preprocessing stage, segments are extracted from aggregated metadata or through the application of heuristics that identify common header phrases such as ``Introduction" or ``Discussion". Input texts are truncated at 2048 tokens for \tecnica[\ACMCRED]{I+D} and 512 tokens for \tecnica[\ACMCRED]{S-n/a} and \tecnica[\ACMCRED]{S-w/a}, leaving 98.6\% and 66\% of texts unchanged, respectively. Reference abstracts are truncated at 512 tokens, a length surpassing that of 98\% of examples. The dataset is randomly partitioned into training (70\%), validation (15\%), and test (15\%) sets. Final metric performances are reported on the test set using models selected based on validation results.

This scientific summarization dataset provides a challenging and realistic benchmark to rigorously evaluate our proposed prompting techniques when integrated with modern transformer architectures.

\subsection{Training}

We perform additional Fine-Tuning on our dataset to integrate prompting capabilities within summarization models. As model capabilities vary, we tune hyperparameters, including learning rates and gradient accumulation steps, on validation results using Ray Tune \cite{raytune}. 

The core training procedure is as follows. In each batch, prompts are formulated per the specified technique outlined in Section \ref{sec:methods:prompting-technique-dimension}. For each model pertaining to Section \ref{sec:methods:model-dimension}, training is executed using a target output corresponding to the prompt's concatenation with the target summary derived from the training example. This concatenation is contingent upon the input text dimension, as described in Section \ref{sec:methods:input-text-dimension}. Training is conducted utilizing teacher forcing, and validation ROUGE is monitored to implement early stopping, preventing overfitting or inefficient resource utilization. During inference, only prompts and input texts are supplied as input.

We compare models trained on our dataset with versus without prompts to quantify the impact of prompting techniques. As a strong baseline, all models are first fine-tuned on the dataset without prompts, referred to as Fine-Tuning. This measures their out-of-the-box summarization capabilities. We then compare Fine-Tuning with our prompting techniques, e.g., \tecnica[\ACMCMAGENTA]{KeyBERT}.

\subsection{Metrics}
\label{sec:results:metrics}

We present the primary summarization results for the test split in Table \ref{tab:main-results}. Table \ref{tab:results-improvement} illustrates the relative improvement of the techniques compared to Fine-Tuning, computed according to Equation \ref{eq:improvement}, where the scores of $\text{V}_{\text{Technique}}$ and $\text{V}_{\text{Fine-Tuning}}$ are calculated using ROUGE metrics \cite{rouge}.

\begin{equation}
    \label{eq:improvement}
    \text{Improvement} = \frac{\text{V}_{\text{Technique}} - \text{V}_{\text{Fine-Tuning}}}{\text{V}_{\text{Fine-Tuning}}}
\end{equation}

In addition to overall performance, we conduct targeted confusion testing by feeding models with prompts extracted from unrelated articles. If prompts improve focusing, the provision of irrelevant prompts should decrease quality compared to proper in-domain prompts. More significant degradation implies better exploitation of prompt information. We analyze confusion testing results to isolate the benefits of prompting.

\begin{table}[htbp]
        \centering
        \begin{adjustbox}{angle=90, scale=0.80}
        \begin{tabular}{|c|c|c|c|c|c|c|c|c|c|c|c|c|c|c|c|}
        \hline
        \rowcolor{gray!40}
        \multicolumn{1}{|c|}{} &
        \multicolumn{3}{c|}{\tecnica[\ACMCBLUE]{BigBirdPegasus}} & \multicolumn{3}{c|}{\tecnica[\ACMCBLUE]{LED}} & \multicolumn{3}{c|}{\tecnica[\ACMCBLUE]{LT5-Base-Local}} & \multicolumn{3}{c|}{\tecnica[\ACMCBLUE]{LT5-Base-ETC}} & \multicolumn{3}{c|}{\tecnica[\ACMCBLUE]{LT5-Large-ETC}} \\
        \hline
        \rowcolor{gray!40}
        & \tecnica[\ACMCRED]{I+D} & \tecnica[\ACMCRED]{S-n/a} & \tecnica[\ACMCRED]{S-w/a} & \tecnica[\ACMCRED]{I+D} & \tecnica[\ACMCRED]{S-n/a} & \tecnica[\ACMCRED]{S-w/a} & \tecnica[\ACMCRED]{I+D} & \tecnica[\ACMCRED]{S-n/a} & \tecnica[\ACMCRED]{S-w/a} & \tecnica[\ACMCRED]{I+D} & \tecnica[\ACMCRED]{S-n/a} & \tecnica[\ACMCRED]{S-w/a} & \tecnica[\ACMCRED]{I+D} & \tecnica[\ACMCRED]{S-n/a} & \tecnica[\ACMCRED]{S-w/a} \\
        \hline
        \hline

	\hline
        \rowcolor{gray!20}
        \multicolumn{16}{|c|}{ROUGE-1} \\
	\hline
	\cellcolor{gray!20} Original &
	0.273 & 0.356 & \cellcolor{green!80} $\mathbf{0.347}$ & 
	\cellcolor{red!20} $0.401$ & \cellcolor{red!20} $0.312$ & \cellcolor{red!20} $0.316$ & 
	\cellcolor{red!20} $0.299$ & \cellcolor{red!20} $0.348$ & \cellcolor{red!20} $0.344$ & 
	\cellcolor{red!20} $0.215$ & \cellcolor{red!20} $0.388$ & \cellcolor{red!20} $0.384$ & 
	0.364 & \cellcolor{red!20} $0.332$ & \cellcolor{red!20} $0.324$ \\
	\hline
	\cellcolor{gray!20} Fine-Tuning &
	\cellcolor{green!50} $0.276$ & 0.345 & \cellcolor{green!50} $0.346$ & 
	0.424 & 0.416 & 0.422 & 
	0.443 & 0.415 & \cellcolor{green!50} $0.443$ & 
	0.443 & 0.411 & 0.419 & 
	\cellcolor{green!80} $\mathbf{0.416}$ & 0.395 & \cellcolor{green!80} $\mathbf{0.392}$ \\
	\hline
	\cellcolor{gray!20} \tecnica[\ACMCMAGENTA]{Keywords} &
	0.241 & \cellcolor{green!80} $\mathbf{0.372}$ & 0.337 & 
	0.428 & 0.455 & 0.467 & 
	0.464 & \cellcolor{green!80} $\mathbf{0.442}$ & \cellcolor{green!80} $\mathbf{0.448}$ & 
	0.464 & \cellcolor{green!50} $0.473$ & 0.465 & 
	\cellcolor{green!50} $0.401$ & 0.395 & 0.343 \\
	\hline
	\cellcolor{gray!20} \tecnica[\ACMCMAGENTA]{KeyBERT} &
	\cellcolor{green!80} $\mathbf{0.299}$ & 0.327 & 0.302 & 
	\cellcolor{green!50} $0.465$ & \cellcolor{green!50} $0.470$ & \cellcolor{green!50} $0.503$ & 
	\cellcolor{green!50} $0.465$ & 0.431 & 0.407 & 
	\cellcolor{green!50} $0.477$ & \cellcolor{green!80} $\mathbf{0.502}$ & \cellcolor{green!80} $\mathbf{0.520}$ & 
	0.384 & \cellcolor{green!50} $0.410$ & 0.357 \\
	\hline
	\cellcolor{gray!20} \tecnica[\ACMCMAGENTA]{MeSH} &
	0.245 & \cellcolor{green!50} $0.364$ & \cellcolor{red!20} $0.296$ & 
	\cellcolor{green!80} $\mathbf{0.473}$ & 0.446 & 0.433 & 
	0.448 & 0.403 & 0.435 & 
	0.470 & 0.457 & 0.435 & 
	\cellcolor{red!20} $0.332$ & 0.399 & 0.379 \\
	\hline
	\cellcolor{gray!20} \tecnica[\ACMCMAGENTA]{TF} &
	\cellcolor{red!20} $0.205$ & \cellcolor{red!20} $0.285$ & 0.316 & 
	0.443 & \cellcolor{green!80} $\mathbf{0.481}$ & 0.492 & 
	\cellcolor{green!80} $\mathbf{0.467}$ & \cellcolor{green!50} $0.433$ & 0.434 & 
	\cellcolor{green!80} $\mathbf{0.480}$ & 0.450 & \cellcolor{green!50} $0.476$ & 
	0.389 & 0.382 & \cellcolor{green!50} $0.380$ \\
	\hline
	\cellcolor{gray!20} \tecnica[\ACMCMAGENTA]{TF-IDF} &
	0.243 & 0.326 & 0.324 & 
	0.441 & 0.453 & \cellcolor{green!80} $\mathbf{0.503}$ & 
	0.464 & 0.409 & 0.437 & 
	0.470 & 0.454 & 0.425 & 
	0.378 & \cellcolor{green!80} $\mathbf{0.411}$ & 0.348 \\
	\hline
	\hline
        \rowcolor{gray!20}
        \multicolumn{16}{|c|}{ROUGE-2} \\
	\hline
	\cellcolor{gray!20} Original &
	0.048 & 0.080 & 0.079 & 
	\cellcolor{red!20} $0.129$ & \cellcolor{red!20} $0.154$ & \cellcolor{red!20} $0.155$ & 
	\cellcolor{red!20} $0.110$ & 0.155 & \cellcolor{red!20} $0.152$ & 
	\cellcolor{red!20} $0.071$ & \cellcolor{red!20} $0.141$ & \cellcolor{red!20} $0.140$ & 
	\cellcolor{green!80} $\mathbf{0.154}$ & 0.138 & 0.133 \\
	\hline
	\cellcolor{gray!20} Fine-Tuning &
	0.050 & 0.081 & 0.083 & 
	0.153 & 0.184 & 0.183 & 
	0.157 & \cellcolor{red!20} $0.142$ & 0.154 & 
	0.183 & 0.188 & 0.175 & 
	0.145 & \cellcolor{red!20} $0.136$ & \cellcolor{red!20} $0.122$ \\
	\hline
	\cellcolor{gray!20} \tecnica[\ACMCMAGENTA]{Keywords} &
	\cellcolor{green!50} $0.052$ & 0.097 & 0.089 & 
	0.155 & \cellcolor{green!50} $0.209$ & 0.218 & 
	0.171 & \cellcolor{green!80} $\mathbf{0.176}$ & \cellcolor{green!80} $\mathbf{0.172}$ & 
	0.181 & \cellcolor{green!50} $0.221$ & 0.216 & 
	\cellcolor{green!50} $0.149$ & 0.141 & 0.135 \\
	\hline
	\cellcolor{gray!20} \tecnica[\ACMCMAGENTA]{KeyBERT} &
	\cellcolor{green!80} $\mathbf{0.056}$ & \cellcolor{green!80} $\mathbf{0.103}$ & \cellcolor{green!50} $0.101$ & 
	\cellcolor{green!80} $\mathbf{0.179}$ & 0.200 & \cellcolor{green!50} $0.222$ & 
	\cellcolor{green!50} $0.172$ & 0.155 & 0.154 & 
	\cellcolor{green!80} $\mathbf{0.200}$ & \cellcolor{green!80} $\mathbf{0.238}$ & \cellcolor{green!80} $\mathbf{0.256}$ & 
	0.144 & \cellcolor{green!50} $0.147$ & 0.139 \\
	\hline
	\cellcolor{gray!20} \tecnica[\ACMCMAGENTA]{MeSH} &
	0.050 & 0.091 & \cellcolor{red!20} $0.070$ & 
	0.165 & 0.200 & 0.195 & 
	0.164 & 0.152 & 0.166 & 
	0.183 & 0.210 & 0.196 & 
	\cellcolor{red!20} $0.123$ & 0.138 & \cellcolor{green!50} $0.142$ \\
	\hline
	\cellcolor{gray!20} \tecnica[\ACMCMAGENTA]{TF} &
	\cellcolor{red!20} $0.032$ & \cellcolor{green!50} $0.101$ & \cellcolor{green!80} $\mathbf{0.105}$ & 
	\cellcolor{green!50} $0.169$ & 0.207 & 0.214 & 
	\cellcolor{green!80} $\mathbf{0.173}$ & \cellcolor{green!50} $0.166$ & 0.167 & 
	\cellcolor{green!50} $0.199$ & 0.177 & \cellcolor{green!50} $0.227$ & 
	0.146 & 0.139 & \cellcolor{green!80} $\mathbf{0.143}$ \\
	\hline
	\cellcolor{gray!20} \tecnica[\ACMCMAGENTA]{TF-IDF} &
	0.045 & \cellcolor{red!20} $0.079$ & 0.073 & 
	0.159 & \cellcolor{green!80} $\mathbf{0.218}$ & \cellcolor{green!80} $\mathbf{0.231}$ & 
	0.172 & 0.159 & \cellcolor{green!50} $0.171$ & 
	0.193 & 0.211 & 0.191 & 
	0.141 & \cellcolor{green!80} $\mathbf{0.156}$ & 0.140 \\
	\hline
	\hline
        \rowcolor{gray!20}
        \multicolumn{16}{|c|}{ROUGE-Lsum} \\
	\hline
	\cellcolor{gray!20} Original &
	\cellcolor{green!80} $\mathbf{0.224}$ & \cellcolor{green!50} $0.301$ & \cellcolor{green!80} $\mathbf{0.295}$ & 
	\cellcolor{red!20} $0.282$ & \cellcolor{red!20} $0.254$ & \cellcolor{red!20} $0.259$ & 
	\cellcolor{red!20} $0.219$ & \cellcolor{red!20} $0.229$ & \cellcolor{red!20} $0.227$ & 
	\cellcolor{red!20} $0.169$ & \cellcolor{red!20} $0.322$ & \cellcolor{red!20} $0.321$ & 
	\cellcolor{red!20} $0.249$ & \cellcolor{red!20} $0.230$ & \cellcolor{red!20} $0.226$ \\
	\hline
	\cellcolor{gray!20} Fine-Tuning &
	0.183 & 0.279 & \cellcolor{green!50} $0.282$ & 
	0.358 & 0.350 & 0.362 & 
	0.333 & 0.350 & \cellcolor{green!50} $0.372$ & 
	0.348 & 0.351 & 0.351 & 
	\cellcolor{green!80} $\mathbf{0.304}$ & 0.334 & \cellcolor{green!80} $\mathbf{0.327}$ \\
	\hline
	\cellcolor{gray!20} \tecnica[\ACMCMAGENTA]{Keywords} &
	0.169 & \cellcolor{green!80} $\mathbf{0.302}$ & 0.278 & 
	0.356 & \cellcolor{green!50} $0.394$ & 0.401 & 
	0.342 & \cellcolor{green!80} $\mathbf{0.374}$ & \cellcolor{green!80} $\mathbf{0.377}$ & 
	0.350 & \cellcolor{green!50} $0.404$ & 0.397 & 
	\cellcolor{green!50} $0.296$ & 0.336 & 0.296 \\
	\hline
	\cellcolor{gray!20} \tecnica[\ACMCMAGENTA]{KeyBERT} &
	\cellcolor{green!50} $0.196$ & 0.286 & 0.262 & 
	\cellcolor{green!80} $\mathbf{0.387}$ & 0.382 & \cellcolor{green!50} $0.416$ & 
	\cellcolor{green!80} $\mathbf{0.345}$ & 0.361 & 0.347 & 
	\cellcolor{green!50} $0.356$ & \cellcolor{green!80} $\mathbf{0.426}$ & \cellcolor{green!80} $\mathbf{0.439}$ & 
	0.291 & \cellcolor{green!50} $0.347$ & 0.309 \\
	\hline
	\cellcolor{gray!20} \tecnica[\ACMCMAGENTA]{MeSH} &
	0.173 & 0.294 & \cellcolor{red!20} $0.246$ & 
	\cellcolor{green!50} $0.385$ & 0.377 & 0.369 & 
	0.330 & 0.343 & 0.365 & 
	0.343 & 0.390 & 0.370 & 
	0.251 & 0.338 & \cellcolor{green!50} $0.327$ \\
	\hline
	\cellcolor{gray!20} \tecnica[\ACMCMAGENTA]{TF} &
	\cellcolor{red!20} $0.148$ & \cellcolor{red!20} $0.240$ & 0.266 & 
	0.373 & \cellcolor{green!80} $\mathbf{0.403}$ & 0.407 & 
	\cellcolor{green!50} $0.344$ & \cellcolor{green!50} $0.365$ & 0.363 & 
	\cellcolor{green!80} $\mathbf{0.357}$ & 0.380 & \cellcolor{green!50} $0.403$ & 
	0.291 & 0.323 & 0.324 \\
	\hline
	\cellcolor{gray!20} \tecnica[\ACMCMAGENTA]{TF-IDF} &
	0.172 & 0.267 & 0.266 & 
	0.366 & 0.392 & \cellcolor{green!80} $\mathbf{0.422}$ & 
	0.341 & 0.350 & 0.367 & 
	0.347 & 0.385 & 0.363 & 
	0.279 & \cellcolor{green!80} $\mathbf{0.350}$ & 0.304 \\
	\hline

        \end{tabular}
        \end{adjustbox}
        \caption{Test results.}
        \label{tab:main-results}
        \end{table}

\begin{table}[htbp]
        \centering
        \begin{adjustbox}{angle=90, scale=0.80}
        \begin{tabular}{|c|c|c|c|c|c|c|c|c|c|c|c|c|c|c|c|}
        \hline
        \rowcolor{gray!40}
        \multicolumn{1}{|c|}{} &
        \multicolumn{3}{c|}{\tecnica[\ACMCBLUE]{BigBirdPegasus}} & \multicolumn{3}{c|}{\tecnica[\ACMCBLUE]{LED}} & \multicolumn{3}{c|}{\tecnica[\ACMCBLUE]{LT5-Base-Local}} & \multicolumn{3}{c|}{\tecnica[\ACMCBLUE]{LT5-Base-ETC}} & \multicolumn{3}{c|}{\tecnica[\ACMCBLUE]{LT5-Large-ETC}} \\
        \hline
        \rowcolor{gray!40}
        & \tecnica[\ACMCRED]{I+D} & \tecnica[\ACMCRED]{S-n/a} & \tecnica[\ACMCRED]{S-w/a} & \tecnica[\ACMCRED]{I+D} & \tecnica[\ACMCRED]{S-n/a} & \tecnica[\ACMCRED]{S-w/a} & \tecnica[\ACMCRED]{I+D} & \tecnica[\ACMCRED]{S-n/a} & \tecnica[\ACMCRED]{S-w/a} & \tecnica[\ACMCRED]{I+D} & \tecnica[\ACMCRED]{S-n/a} & \tecnica[\ACMCRED]{S-w/a} & \tecnica[\ACMCRED]{I+D} & \tecnica[\ACMCRED]{S-n/a} & \tecnica[\ACMCRED]{S-w/a} \\
        \hline
        \hline

	\hline
        \rowcolor{gray!20}
        \multicolumn{16}{|c|}{ROUGE-1} \\
	\hline
	\cellcolor{gray!20} \tecnica[\ACMCMAGENTA]{Keywords} &
	\cellcolor{red!12} $-0.126$ & \cellcolor{green!8} $0.080$ & \cellcolor{red!2} $-0.027$ & 
	\cellcolor{green!1} $0.010$ & \cellcolor{green!9} $0.093$ & \cellcolor{green!10} $0.105$ & 
	\cellcolor{green!4} $0.046$ & \cellcolor{green!6} $0.064$ & \cellcolor{green!1} $0.011$ & 
	\cellcolor{green!4} $0.048$ & \cellcolor{green!15} $0.151$ & \cellcolor{green!10} $0.110$ & 
	\cellcolor{red!3} $-0.037$ & \cellcolor{red!0} $-0.000$ & \cellcolor{red!12} $-0.124$ \\
	\hline
	\cellcolor{gray!20} \tecnica[\ACMCMAGENTA]{KeyBERT} &
	\cellcolor{green!8} $0.086$ & \cellcolor{red!5} $-0.052$ & \cellcolor{red!12} $-0.128$ & 
	\cellcolor{green!9} $0.097$ & \cellcolor{green!12} $0.130$ & \cellcolor{green!19} $0.192$ & 
	\cellcolor{green!4} $0.048$ & \cellcolor{green!3} $0.037$ & \cellcolor{red!8} $-0.081$ & 
	\cellcolor{green!7} $0.077$ & \cellcolor{green!22} $0.222$ & \cellcolor{green!24} $0.241$ & 
	\cellcolor{red!7} $-0.078$ & \cellcolor{green!3} $0.039$ & \cellcolor{red!8} $-0.088$ \\
	\hline
	\cellcolor{gray!20} \tecnica[\ACMCMAGENTA]{MeSH} &
	\cellcolor{red!11} $-0.110$ & \cellcolor{green!5} $0.057$ & \cellcolor{red!14} $-0.144$ & 
	\cellcolor{green!11} $0.116$ & \cellcolor{green!7} $0.072$ & \cellcolor{green!2} $0.025$ & 
	\cellcolor{green!1} $0.011$ & \cellcolor{red!2} $-0.029$ & \cellcolor{red!1} $-0.020$ & 
	\cellcolor{green!6} $0.062$ & \cellcolor{green!11} $0.112$ & \cellcolor{green!3} $0.040$ & 
	\cellcolor{red!20} $-0.201$ & \cellcolor{green!1} $0.011$ & \cellcolor{red!3} $-0.032$ \\
	\hline
	\cellcolor{gray!20} \tecnica[\ACMCMAGENTA]{TF} &
	\cellcolor{red!25} $-0.256$ & \cellcolor{red!17} $-0.173$ & \cellcolor{red!8} $-0.085$ & 
	\cellcolor{green!4} $0.044$ & \cellcolor{green!15} $0.156$ & \cellcolor{green!16} $0.165$ & 
	\cellcolor{green!5} $0.053$ & \cellcolor{green!4} $0.043$ & \cellcolor{red!2} $-0.021$ & 
	\cellcolor{green!8} $0.084$ & \cellcolor{green!9} $0.096$ & \cellcolor{green!13} $0.136$ & 
	\cellcolor{red!6} $-0.065$ & \cellcolor{red!3} $-0.033$ & \cellcolor{red!3} $-0.031$ \\
	\hline
	\cellcolor{gray!20} \tecnica[\ACMCMAGENTA]{TF-IDF} &
	\cellcolor{red!11} $-0.117$ & \cellcolor{red!5} $-0.054$ & \cellcolor{red!6} $-0.064$ & 
	\cellcolor{green!3} $0.040$ & \cellcolor{green!8} $0.088$ & \cellcolor{green!19} $0.192$ & 
	\cellcolor{green!4} $0.046$ & \cellcolor{red!1} $-0.015$ & \cellcolor{red!1} $-0.015$ & 
	\cellcolor{green!6} $0.060$ & \cellcolor{green!10} $0.105$ & \cellcolor{green!1} $0.016$ & 
	\cellcolor{red!9} $-0.092$ & \cellcolor{green!4} $0.042$ & \cellcolor{red!11} $-0.111$ \\
	\hline
	\hline
        \rowcolor{gray!20}
        \multicolumn{16}{|c|}{ROUGE-2} \\
	\hline
	\cellcolor{gray!20} \tecnica[\ACMCMAGENTA]{Keywords} &
	\cellcolor{green!3} $0.036$ & \cellcolor{green!19} $0.191$ & \cellcolor{green!6} $0.065$ & 
	\cellcolor{green!1} $0.016$ & \cellcolor{green!13} $0.132$ & \cellcolor{green!19} $0.192$ & 
	\cellcolor{green!8} $0.089$ & \cellcolor{green!23} $0.238$ & \cellcolor{green!11} $0.118$ & 
	\cellcolor{red!0} $-0.009$ & \cellcolor{green!17} $0.176$ & \cellcolor{green!23} $0.232$ & 
	\cellcolor{green!2} $0.028$ & \cellcolor{green!3} $0.037$ & \cellcolor{green!10} $0.105$ \\
	\hline
	\cellcolor{gray!20} \tecnica[\ACMCMAGENTA]{KeyBERT} &
	\cellcolor{green!12} $0.120$ & \cellcolor{green!26} $0.260$ & \cellcolor{green!21} $0.217$ & 
	\cellcolor{green!17} $0.170$ & \cellcolor{green!8} $0.086$ & \cellcolor{green!21} $0.211$ & 
	\cellcolor{green!9} $0.096$ & \cellcolor{green!9} $0.091$ & \cellcolor{green!0} $0.002$ & 
	\cellcolor{green!9} $0.095$ & \cellcolor{green!26} $0.268$ & \cellcolor{green!46} $0.462$ & 
	\cellcolor{red!0} $-0.010$ & \cellcolor{green!8} $0.085$ & \cellcolor{green!13} $0.137$ \\
	\hline
	\cellcolor{gray!20} \tecnica[\ACMCMAGENTA]{MeSH} &
	\cellcolor{red!0} $-0.010$ & \cellcolor{green!11} $0.117$ & \cellcolor{red!16} $-0.164$ & 
	\cellcolor{green!7} $0.076$ & \cellcolor{green!8} $0.084$ & \cellcolor{green!6} $0.064$ & 
	\cellcolor{green!4} $0.044$ & \cellcolor{green!6} $0.065$ & \cellcolor{green!8} $0.084$ & 
	\cellcolor{green!0} $0.003$ & \cellcolor{green!11} $0.114$ & \cellcolor{green!11} $0.120$ & 
	\cellcolor{red!15} $-0.155$ & \cellcolor{green!2} $0.021$ & \cellcolor{green!16} $0.161$ \\
	\hline
	\cellcolor{gray!20} \tecnica[\ACMCMAGENTA]{TF} &
	\cellcolor{red!35} $-0.355$ & \cellcolor{green!24} $0.243$ & \cellcolor{green!26} $0.266$ & 
	\cellcolor{green!10} $0.107$ & \cellcolor{green!12} $0.125$ & \cellcolor{green!17} $0.172$ & 
	\cellcolor{green!10} $0.106$ & \cellcolor{green!16} $0.163$ & \cellcolor{green!8} $0.085$ & 
	\cellcolor{green!8} $0.089$ & \cellcolor{red!5} $-0.059$ & \cellcolor{green!29} $0.294$ & 
	\cellcolor{green!0} $0.007$ & \cellcolor{green!2} $0.025$ & \cellcolor{green!16} $0.169$ \\
	\hline
	\cellcolor{gray!20} \tecnica[\ACMCMAGENTA]{TF-IDF} &
	\cellcolor{red!9} $-0.094$ & \cellcolor{red!2} $-0.029$ & \cellcolor{red!12} $-0.124$ & 
	\cellcolor{green!3} $0.038$ & \cellcolor{green!18} $0.181$ & \cellcolor{green!26} $0.260$ & 
	\cellcolor{green!9} $0.096$ & \cellcolor{green!11} $0.119$ & \cellcolor{green!11} $0.112$ & 
	\cellcolor{green!5} $0.058$ & \cellcolor{green!12} $0.123$ & \cellcolor{green!8} $0.090$ & 
	\cellcolor{red!2} $-0.026$ & \cellcolor{green!15} $0.152$ & \cellcolor{green!14} $0.149$ \\
	\hline
	\hline
        \rowcolor{gray!20}
        \multicolumn{16}{|c|}{ROUGE-Lsum} \\
	\hline
	\cellcolor{gray!20} \tecnica[\ACMCMAGENTA]{Keywords} &
	\cellcolor{red!7} $-0.077$ & \cellcolor{green!8} $0.081$ & \cellcolor{red!1} $-0.013$ & 
	\cellcolor{red!0} $-0.004$ & \cellcolor{green!12} $0.125$ & \cellcolor{green!10} $0.109$ & 
	\cellcolor{green!2} $0.028$ & \cellcolor{green!6} $0.067$ & \cellcolor{green!1} $0.014$ & 
	\cellcolor{green!0} $0.005$ & \cellcolor{green!15} $0.153$ & \cellcolor{green!13} $0.131$ & 
	\cellcolor{red!2} $-0.028$ & \cellcolor{green!0} $0.005$ & \cellcolor{red!9} $-0.095$ \\
	\hline
	\cellcolor{gray!20} \tecnica[\ACMCMAGENTA]{KeyBERT} &
	\cellcolor{green!7} $0.074$ & \cellcolor{green!2} $0.024$ & \cellcolor{red!7} $-0.070$ & 
	\cellcolor{green!8} $0.081$ & \cellcolor{green!9} $0.090$ & \cellcolor{green!15} $0.151$ & 
	\cellcolor{green!3} $0.036$ & \cellcolor{green!3} $0.032$ & \cellcolor{red!6} $-0.067$ & 
	\cellcolor{green!2} $0.022$ & \cellcolor{green!21} $0.215$ & \cellcolor{green!25} $0.251$ & 
	\cellcolor{red!4} $-0.045$ & \cellcolor{green!3} $0.037$ & \cellcolor{red!5} $-0.056$ \\
	\hline
	\cellcolor{gray!20} \tecnica[\ACMCMAGENTA]{MeSH} &
	\cellcolor{red!5} $-0.054$ & \cellcolor{green!5} $0.052$ & \cellcolor{red!12} $-0.128$ & 
	\cellcolor{green!7} $0.076$ & \cellcolor{green!7} $0.078$ & \cellcolor{green!2} $0.020$ & 
	\cellcolor{red!0} $-0.007$ & \cellcolor{red!2} $-0.021$ & \cellcolor{red!1} $-0.020$ & 
	\cellcolor{red!1} $-0.016$ & \cellcolor{green!11} $0.113$ & \cellcolor{green!5} $0.053$ & 
	\cellcolor{red!17} $-0.175$ & \cellcolor{green!1} $0.012$ & \cellcolor{red!0} $-0.001$ \\
	\hline
	\cellcolor{gray!20} \tecnica[\ACMCMAGENTA]{TF} &
	\cellcolor{red!18} $-0.189$ & \cellcolor{red!13} $-0.139$ & \cellcolor{red!5} $-0.055$ & 
	\cellcolor{green!4} $0.041$ & \cellcolor{green!15} $0.151$ & \cellcolor{green!12} $0.125$ & 
	\cellcolor{green!3} $0.032$ & \cellcolor{green!4} $0.043$ & \cellcolor{red!2} $-0.025$ & 
	\cellcolor{green!2} $0.027$ & \cellcolor{green!8} $0.082$ & \cellcolor{green!14} $0.147$ & 
	\cellcolor{red!4} $-0.043$ & \cellcolor{red!3} $-0.034$ & \cellcolor{red!0} $-0.010$ \\
	\hline
	\cellcolor{gray!20} \tecnica[\ACMCMAGENTA]{TF-IDF} &
	\cellcolor{red!6} $-0.060$ & \cellcolor{red!4} $-0.043$ & \cellcolor{red!5} $-0.057$ & 
	\cellcolor{green!2} $0.023$ & \cellcolor{green!11} $0.119$ & \cellcolor{green!16} $0.166$ & 
	\cellcolor{green!2} $0.024$ & \cellcolor{red!0} $-0.001$ & \cellcolor{red!1} $-0.012$ & 
	\cellcolor{red!0} $-0.003$ & \cellcolor{green!9} $0.098$ & \cellcolor{green!3} $0.033$ & 
	\cellcolor{red!8} $-0.083$ & \cellcolor{green!4} $0.046$ & \cellcolor{red!7} $-0.071$ \\
	\hline

        \end{tabular}
        \end{adjustbox}
        \caption{Improvements of the techniques presented, in relation to Fine-Tuning.}
        \label{tab:results-improvement}
        \end{table}

\begin{table}[htbp]
        \centering
        \begin{adjustbox}{angle=90, scale=0.80}
        \begin{tabular}{|c|c|c|c|c|c|c|c|c|c|c|c|c|c|c|c|}
        \hline
        \rowcolor{gray!40}
        \multicolumn{1}{|c|}{} &
        \multicolumn{3}{c|}{\tecnica[\ACMCBLUE]{BigBirdPegasus}} & \multicolumn{3}{c|}{\tecnica[\ACMCBLUE]{LED}} & \multicolumn{3}{c|}{\tecnica[\ACMCBLUE]{LT5-Base-Local}} & \multicolumn{3}{c|}{\tecnica[\ACMCBLUE]{LT5-Base-ETC}} & \multicolumn{3}{c|}{\tecnica[\ACMCBLUE]{LT5-Large-ETC}} \\
        \hline
        \rowcolor{gray!40}
        & \tecnica[\ACMCRED]{I+D} & \tecnica[\ACMCRED]{S-n/a} & \tecnica[\ACMCRED]{S-w/a} & \tecnica[\ACMCRED]{I+D} & \tecnica[\ACMCRED]{S-n/a} & \tecnica[\ACMCRED]{S-w/a} & \tecnica[\ACMCRED]{I+D} & \tecnica[\ACMCRED]{S-n/a} & \tecnica[\ACMCRED]{S-w/a} & \tecnica[\ACMCRED]{I+D} & \tecnica[\ACMCRED]{S-n/a} & \tecnica[\ACMCRED]{S-w/a} & \tecnica[\ACMCRED]{I+D} & \tecnica[\ACMCRED]{S-n/a} & \tecnica[\ACMCRED]{S-w/a} \\
        \hline
        \hline

	\hline
        \rowcolor{gray!20}
        \multicolumn{16}{|c|}{ROUGE-1} \\
	\hline
	\cellcolor{gray!20} \tecnica[\ACMCMAGENTA]{Keywords} &
	\cellcolor{green!50} $0.255$ & \cellcolor{green!80} $\mathbf{0.360}$ & \cellcolor{green!50} $0.334$ & 
	0.275 & 0.255 & 0.258 & 
	\cellcolor{green!80} $\mathbf{0.440}$ & \cellcolor{green!80} $\mathbf{0.355}$ & \cellcolor{green!80} $\mathbf{0.370}$ & 
	\cellcolor{red!20} $0.406$ & \cellcolor{green!80} $\mathbf{0.380}$ & \cellcolor{green!50} $0.348$ & 
	\cellcolor{green!50} $0.390$ & 0.381 & 0.353 \\
	\hline
	\cellcolor{gray!20} \tecnica[\ACMCMAGENTA]{KeyBERT} &
	\cellcolor{green!80} $\mathbf{0.301}$ & 0.316 & \cellcolor{red!20} $0.286$ & 
	\cellcolor{green!80} $\mathbf{0.282}$ & \cellcolor{red!20} $0.255$ & \cellcolor{red!20} $0.252$ & 
	\cellcolor{green!50} $0.413$ & \cellcolor{green!50} $0.346$ & \cellcolor{red!20} $0.313$ & 
	0.413 & \cellcolor{green!50} $0.358$ & \cellcolor{green!80} $\mathbf{0.371}$ & 
	\cellcolor{green!80} $\mathbf{0.401}$ & 0.381 & \cellcolor{green!50} $0.357$ \\
	\hline
	\cellcolor{gray!20} \tecnica[\ACMCMAGENTA]{MeSH} &
	0.232 & \cellcolor{green!50} $0.350$ & 0.296 & 
	\cellcolor{red!20} $0.273$ & \cellcolor{green!80} $\mathbf{0.278}$ & \cellcolor{green!80} $\mathbf{0.275}$ & 
	0.410 & 0.309 & 0.347 & 
	\cellcolor{green!80} $\mathbf{0.446}$ & 0.346 & \cellcolor{red!20} $0.310$ & 
	\cellcolor{red!20} $0.344$ & \cellcolor{green!50} $0.386$ & 0.356 \\
	\hline
	\cellcolor{gray!20} \tecnica[\ACMCMAGENTA]{TF} &
	\cellcolor{red!20} $0.219$ & \cellcolor{red!20} $0.286$ & \cellcolor{green!80} $\mathbf{0.336}$ & 
	0.277 & \cellcolor{green!50} $0.257$ & 0.260 & 
	\cellcolor{red!20} $0.405$ & 0.339 & \cellcolor{green!50} $0.349$ & 
	0.415 & \cellcolor{red!20} $0.311$ & 0.321 & 
	0.385 & \cellcolor{red!20} $0.370$ & \cellcolor{green!80} $\mathbf{0.365}$ \\
	\hline
	\cellcolor{gray!20} \tecnica[\ACMCMAGENTA]{TF-IDF} &
	0.239 & 0.322 & 0.317 & 
	\cellcolor{green!50} $0.280$ & 0.256 & \cellcolor{green!50} $0.267$ & 
	0.408 & \cellcolor{red!20} $0.294$ & 0.345 & 
	\cellcolor{green!50} $0.415$ & 0.341 & 0.320 & 
	0.354 & \cellcolor{green!80} $\mathbf{0.392}$ & \cellcolor{red!20} $0.332$ \\
	\hline
	\hline
        \rowcolor{gray!20}
        \multicolumn{16}{|c|}{ROUGE-2} \\
	\hline
	\cellcolor{gray!20} \tecnica[\ACMCMAGENTA]{Keywords} &
	\cellcolor{green!50} $0.055$ & 0.086 & 0.078 & 
	\cellcolor{red!20} $0.041$ & 0.056 & \cellcolor{red!20} $0.056$ & 
	\cellcolor{green!80} $\mathbf{0.143}$ & \cellcolor{green!80} $\mathbf{0.131}$ & \cellcolor{green!80} $\mathbf{0.137}$ & 
	\cellcolor{red!20} $0.152$ & \cellcolor{green!80} $\mathbf{0.158}$ & \cellcolor{green!50} $0.153$ & 
	\cellcolor{green!50} $0.146$ & \cellcolor{green!50} $0.133$ & \cellcolor{green!50} $0.132$ \\
	\hline
	\cellcolor{gray!20} \tecnica[\ACMCMAGENTA]{KeyBERT} &
	\cellcolor{green!80} $\mathbf{0.058}$ & \cellcolor{green!80} $\mathbf{0.123}$ & \cellcolor{green!50} $0.112$ & 
	\cellcolor{green!50} $0.042$ & \cellcolor{green!80} $\mathbf{0.059}$ & 0.056 & 
	0.133 & 0.123 & \cellcolor{red!20} $0.111$ & 
	0.154 & \cellcolor{green!50} $0.157$ & \cellcolor{green!80} $\mathbf{0.159}$ & 
	\cellcolor{green!80} $\mathbf{0.153}$ & 0.126 & \cellcolor{green!80} $\mathbf{0.137}$ \\
	\hline
	\cellcolor{gray!20} \tecnica[\ACMCMAGENTA]{MeSH} &
	0.040 & 0.074 & \cellcolor{red!20} $0.057$ & 
	0.041 & \cellcolor{green!50} $0.059$ & \cellcolor{green!50} $0.058$ & 
	0.131 & 0.110 & 0.124 & 
	\cellcolor{green!80} $\mathbf{0.178}$ & 0.152 & \cellcolor{red!20} $0.130$ & 
	\cellcolor{red!20} $0.124$ & 0.126 & \cellcolor{red!20} $0.114$ \\
	\hline
	\cellcolor{gray!20} \tecnica[\ACMCMAGENTA]{TF} &
	\cellcolor{red!20} $0.031$ & \cellcolor{green!50} $0.095$ & \cellcolor{green!80} $\mathbf{0.115}$ & 
	0.042 & 0.058 & 0.057 & 
	\cellcolor{red!20} $0.122$ & \cellcolor{green!50} $0.125$ & 0.127 & 
	\cellcolor{green!50} $0.159$ & \cellcolor{red!20} $0.116$ & 0.150 & 
	0.141 & \cellcolor{red!20} $0.122$ & 0.127 \\
	\hline
	\cellcolor{gray!20} \tecnica[\ACMCMAGENTA]{TF-IDF} &
	0.045 & \cellcolor{red!20} $0.066$ & 0.065 & 
	\cellcolor{green!80} $\mathbf{0.043}$ & \cellcolor{red!20} $0.054$ & \cellcolor{green!80} $\mathbf{0.059}$ & 
	\cellcolor{green!50} $0.142$ & \cellcolor{red!20} $0.101$ & \cellcolor{green!50} $0.128$ & 
	0.159 & 0.151 & 0.136 & 
	0.126 & \cellcolor{green!80} $\mathbf{0.138}$ & 0.118 \\
	\hline
	\hline
        \rowcolor{gray!20}
        \multicolumn{16}{|c|}{ROUGE-Lsum} \\
	\hline
	\cellcolor{gray!20} \tecnica[\ACMCMAGENTA]{Keywords} &
	\cellcolor{green!50} $0.173$ & \cellcolor{green!80} $\mathbf{0.289}$ & \cellcolor{green!50} $0.267$ & 
	\cellcolor{red!20} $0.149$ & \cellcolor{red!20} $0.178$ & 0.180 & 
	\cellcolor{green!80} $\mathbf{0.319}$ & \cellcolor{green!80} $\mathbf{0.304}$ & \cellcolor{green!80} $\mathbf{0.314}$ & 
	0.306 & \cellcolor{green!80} $\mathbf{0.327}$ & \cellcolor{green!50} $0.304$ & 
	\cellcolor{green!50} $0.293$ & \cellcolor{green!50} $0.325$ & 0.304 \\
	\hline
	\cellcolor{gray!20} \tecnica[\ACMCMAGENTA]{KeyBERT} &
	\cellcolor{green!80} $\mathbf{0.196}$ & 0.274 & 0.249 & 
	\cellcolor{green!80} $\mathbf{0.153}$ & \cellcolor{green!50} $0.180$ & \cellcolor{red!20} $0.175$ & 
	\cellcolor{green!50} $0.307$ & \cellcolor{green!50} $0.298$ & \cellcolor{red!20} $0.271$ & 
	\cellcolor{red!20} $0.301$ & \cellcolor{green!50} $0.313$ & \cellcolor{green!80} $\mathbf{0.321}$ & 
	\cellcolor{green!80} $\mathbf{0.304}$ & 0.319 & \cellcolor{green!50} $0.309$ \\
	\hline
	\cellcolor{gray!20} \tecnica[\ACMCMAGENTA]{MeSH} &
	0.158 & \cellcolor{green!50} $0.278$ & \cellcolor{red!20} $0.239$ & 
	0.149 & \cellcolor{green!80} $\mathbf{0.193}$ & \cellcolor{green!80} $\mathbf{0.190}$ & 
	0.295 & 0.272 & 0.299 & 
	\cellcolor{green!80} $\mathbf{0.326}$ & 0.301 & \cellcolor{red!20} $0.268$ & 
	0.262 & 0.324 & 0.301 \\
	\hline
	\cellcolor{gray!20} \tecnica[\ACMCMAGENTA]{TF} &
	\cellcolor{red!20} $0.152$ & \cellcolor{red!20} $0.241$ & \cellcolor{green!80} $\mathbf{0.281}$ & 
	0.151 & 0.179 & 0.183 & 
	\cellcolor{red!20} $0.291$ & 0.289 & \cellcolor{green!50} $0.300$ & 
	\cellcolor{green!50} $0.313$ & \cellcolor{red!20} $0.269$ & 0.287 & 
	0.286 & \cellcolor{red!20} $0.308$ & \cellcolor{green!80} $\mathbf{0.313}$ \\
	\hline
	\cellcolor{gray!20} \tecnica[\ACMCMAGENTA]{TF-IDF} &
	0.164 & 0.260 & 0.254 & 
	\cellcolor{green!50} $0.152$ & 0.178 & \cellcolor{green!50} $0.186$ & 
	0.299 & \cellcolor{red!20} $0.250$ & 0.296 & 
	0.301 & 0.297 & 0.282 & 
	\cellcolor{red!20} $0.262$ & \cellcolor{green!80} $\mathbf{0.330}$ & \cellcolor{red!20} $0.287$ \\
	\hline

        \end{tabular}
        \end{adjustbox}
        \caption{Confusion tests results.}
        \label{tab:results-confusion}
        \end{table}

\begin{table}[htbp]
        \centering
        \begin{adjustbox}{angle=90, scale=0.80}
        \begin{tabular}{|c|c|c|c|c|c|c|c|c|c|c|c|c|c|c|c|}
        \hline
        \rowcolor{gray!40}
        \multicolumn{1}{|c|}{} &
        \multicolumn{3}{c|}{\tecnica[\ACMCBLUE]{BigBirdPegasus}} & \multicolumn{3}{c|}{\tecnica[\ACMCBLUE]{LED}} & \multicolumn{3}{c|}{\tecnica[\ACMCBLUE]{LT5-Base-Local}} & \multicolumn{3}{c|}{\tecnica[\ACMCBLUE]{LT5-Base-ETC}} & \multicolumn{3}{c|}{\tecnica[\ACMCBLUE]{LT5-Large-ETC}} \\
        \hline
        \rowcolor{gray!40}
        & \tecnica[\ACMCRED]{I+D} & \tecnica[\ACMCRED]{S-n/a} & \tecnica[\ACMCRED]{S-w/a} & \tecnica[\ACMCRED]{I+D} & \tecnica[\ACMCRED]{S-n/a} & \tecnica[\ACMCRED]{S-w/a} & \tecnica[\ACMCRED]{I+D} & \tecnica[\ACMCRED]{S-n/a} & \tecnica[\ACMCRED]{S-w/a} & \tecnica[\ACMCRED]{I+D} & \tecnica[\ACMCRED]{S-n/a} & \tecnica[\ACMCRED]{S-w/a} & \tecnica[\ACMCRED]{I+D} & \tecnica[\ACMCRED]{S-n/a} & \tecnica[\ACMCRED]{S-w/a} \\
        \hline
        \hline

	\hline
        \rowcolor{gray!20}
        \multicolumn{16}{|c|}{ROUGE-1} \\
	\hline
	\cellcolor{gray!20} \tecnica[\ACMCMAGENTA]{Keywords} &
	\cellcolor{green!5} $0.060$ & \cellcolor{red!3} $-0.033$ & \cellcolor{red!0} $-0.009$ & 
	\cellcolor{red!35} $-0.357$ & \cellcolor{red!43} $-0.440$ & \cellcolor{red!44} $-0.448$ & 
	\cellcolor{red!5} $-0.050$ & \cellcolor{red!19} $-0.196$ & \cellcolor{red!17} $-0.175$ & 
	\cellcolor{red!12} $-0.125$ & \cellcolor{red!19} $-0.195$ & \cellcolor{red!25} $-0.252$ & 
	\cellcolor{red!2} $-0.027$ & \cellcolor{red!3} $-0.033$ & \cellcolor{green!2} $0.029$ \\
	\hline
	\cellcolor{gray!20} \tecnica[\ACMCMAGENTA]{KeyBERT} &
	\cellcolor{green!0} $0.006$ & \cellcolor{red!3} $-0.033$ & \cellcolor{red!4} $-0.050$ & 
	\cellcolor{red!39} $-0.395$ & \cellcolor{red!45} $-0.458$ & \cellcolor{red!49} $-0.500$ & 
	\cellcolor{red!11} $-0.110$ & \cellcolor{red!19} $-0.197$ & \cellcolor{red!23} $-0.233$ & 
	\cellcolor{red!13} $-0.134$ & \cellcolor{red!28} $-0.287$ & \cellcolor{red!28} $-0.286$ & 
	\cellcolor{green!4} $0.046$ & \cellcolor{red!7} $-0.071$ & \cellcolor{red!0} $-0.002$ \\
	\hline
	\cellcolor{gray!20} \tecnica[\ACMCMAGENTA]{MeSH} &
	\cellcolor{red!5} $-0.055$ & \cellcolor{red!3} $-0.039$ & \cellcolor{red!0} $-0.001$ & 
	\cellcolor{red!42} $-0.424$ & \cellcolor{red!37} $-0.378$ & \cellcolor{red!36} $-0.365$ & 
	\cellcolor{red!8} $-0.085$ & \cellcolor{red!23} $-0.233$ & \cellcolor{red!20} $-0.201$ & 
	\cellcolor{red!5} $-0.052$ & \cellcolor{red!24} $-0.242$ & \cellcolor{red!28} $-0.287$ & 
	\cellcolor{green!3} $0.034$ & \cellcolor{red!3} $-0.033$ & \cellcolor{red!6} $-0.063$ \\
	\hline
	\cellcolor{gray!20} \tecnica[\ACMCMAGENTA]{TF} &
	\cellcolor{green!6} $0.069$ & \cellcolor{green!0} $0.003$ & \cellcolor{green!6} $0.061$ & 
	\cellcolor{red!37} $-0.375$ & \cellcolor{red!46} $-0.465$ & \cellcolor{red!47} $-0.471$ & 
	\cellcolor{red!13} $-0.133$ & \cellcolor{red!21} $-0.217$ & \cellcolor{red!19} $-0.196$ & 
	\cellcolor{red!13} $-0.137$ & \cellcolor{red!30} $-0.308$ & \cellcolor{red!32} $-0.324$ & 
	\cellcolor{red!0} $-0.009$ & \cellcolor{red!3} $-0.031$ & \cellcolor{red!3} $-0.039$ \\
	\hline
	\cellcolor{gray!20} \tecnica[\ACMCMAGENTA]{TF-IDF} &
	\cellcolor{red!1} $-0.019$ & \cellcolor{red!1} $-0.013$ & \cellcolor{red!2} $-0.021$ & 
	\cellcolor{red!36} $-0.365$ & \cellcolor{red!43} $-0.436$ & \cellcolor{red!47} $-0.471$ & 
	\cellcolor{red!11} $-0.119$ & \cellcolor{red!28} $-0.281$ & \cellcolor{red!21} $-0.211$ & 
	\cellcolor{red!11} $-0.117$ & \cellcolor{red!24} $-0.248$ & \cellcolor{red!24} $-0.248$ & 
	\cellcolor{red!6} $-0.064$ & \cellcolor{red!4} $-0.047$ & \cellcolor{red!4} $-0.048$ \\
	\hline
	\hline
        \rowcolor{gray!20}
        \multicolumn{16}{|c|}{ROUGE-2} \\
	\hline
	\cellcolor{gray!20} \tecnica[\ACMCMAGENTA]{Keywords} &
	\cellcolor{green!5} $0.054$ & \cellcolor{red!10} $-0.108$ & \cellcolor{red!11} $-0.116$ & 
	\cellcolor{red!73} $-0.739$ & \cellcolor{red!72} $-0.730$ & \cellcolor{red!74} $-0.743$ & 
	\cellcolor{red!16} $-0.164$ & \cellcolor{red!25} $-0.258$ & \cellcolor{red!20} $-0.200$ & 
	\cellcolor{red!15} $-0.159$ & \cellcolor{red!28} $-0.284$ & \cellcolor{red!29} $-0.291$ & 
	\cellcolor{red!1} $-0.019$ & \cellcolor{red!5} $-0.052$ & \cellcolor{red!2} $-0.022$ \\
	\hline
	\cellcolor{gray!20} \tecnica[\ACMCMAGENTA]{KeyBERT} &
	\cellcolor{green!2} $0.023$ & \cellcolor{green!20} $0.202$ & \cellcolor{green!10} $0.106$ & 
	\cellcolor{red!76} $-0.764$ & \cellcolor{red!70} $-0.705$ & \cellcolor{red!74} $-0.746$ & 
	\cellcolor{red!22} $-0.227$ & \cellcolor{red!20} $-0.207$ & \cellcolor{red!27} $-0.280$ & 
	\cellcolor{red!23} $-0.232$ & \cellcolor{red!34} $-0.342$ & \cellcolor{red!37} $-0.378$ & 
	\cellcolor{green!6} $0.063$ & \cellcolor{red!14} $-0.145$ & \cellcolor{red!1} $-0.014$ \\
	\hline
	\cellcolor{gray!20} \tecnica[\ACMCMAGENTA]{MeSH} &
	\cellcolor{red!18} $-0.189$ & \cellcolor{red!18} $-0.187$ & \cellcolor{red!18} $-0.183$ & 
	\cellcolor{red!75} $-0.752$ & \cellcolor{red!70} $-0.706$ & \cellcolor{red!70} $-0.704$ & 
	\cellcolor{red!20} $-0.201$ & \cellcolor{red!27} $-0.277$ & \cellcolor{red!25} $-0.255$ & 
	\cellcolor{red!2} $-0.027$ & \cellcolor{red!27} $-0.275$ & \cellcolor{red!33} $-0.338$ & 
	\cellcolor{green!0} $0.007$ & \cellcolor{red!9} $-0.091$ & \cellcolor{red!19} $-0.194$ \\
	\hline
	\cellcolor{gray!20} \tecnica[\ACMCMAGENTA]{TF} &
	\cellcolor{red!4} $-0.042$ & \cellcolor{red!6} $-0.062$ & \cellcolor{green!9} $0.091$ & 
	\cellcolor{red!75} $-0.751$ & \cellcolor{red!71} $-0.719$ & \cellcolor{red!73} $-0.736$ & 
	\cellcolor{red!29} $-0.294$ & \cellcolor{red!24} $-0.246$ & \cellcolor{red!23} $-0.238$ & 
	\cellcolor{red!19} $-0.199$ & \cellcolor{red!34} $-0.344$ & \cellcolor{red!33} $-0.338$ & 
	\cellcolor{red!3} $-0.035$ & \cellcolor{red!12} $-0.121$ & \cellcolor{red!11} $-0.114$ \\
	\hline
	\cellcolor{gray!20} \tecnica[\ACMCMAGENTA]{TF-IDF} &
	\cellcolor{red!0} $-0.004$ & \cellcolor{red!16} $-0.164$ & \cellcolor{red!11} $-0.114$ & 
	\cellcolor{red!72} $-0.729$ & \cellcolor{red!75} $-0.750$ & \cellcolor{red!74} $-0.744$ & 
	\cellcolor{red!17} $-0.176$ & \cellcolor{red!36} $-0.366$ & \cellcolor{red!25} $-0.252$ & 
	\cellcolor{red!17} $-0.179$ & \cellcolor{red!28} $-0.283$ & \cellcolor{red!28} $-0.286$ & 
	\cellcolor{red!11} $-0.110$ & \cellcolor{red!11} $-0.120$ & \cellcolor{red!16} $-0.162$ \\
	\hline
	\hline
        \rowcolor{gray!20}
        \multicolumn{16}{|c|}{ROUGE-Lsum} \\
	\hline
	\cellcolor{gray!20} \tecnica[\ACMCMAGENTA]{Keywords} &
	\cellcolor{green!2} $0.028$ & \cellcolor{red!4} $-0.042$ & \cellcolor{red!3} $-0.039$ & 
	\cellcolor{red!58} $-0.583$ & \cellcolor{red!54} $-0.549$ & \cellcolor{red!55} $-0.552$ & 
	\cellcolor{red!6} $-0.068$ & \cellcolor{red!18} $-0.186$ & \cellcolor{red!16} $-0.166$ & 
	\cellcolor{red!12} $-0.126$ & \cellcolor{red!19} $-0.192$ & \cellcolor{red!23} $-0.234$ & 
	\cellcolor{red!0} $-0.008$ & \cellcolor{red!3} $-0.032$ & \cellcolor{green!2} $0.026$ \\
	\hline
	\cellcolor{gray!20} \tecnica[\ACMCMAGENTA]{KeyBERT} &
	\cellcolor{red!0} $-0.002$ & \cellcolor{red!4} $-0.041$ & \cellcolor{red!4} $-0.049$ & 
	\cellcolor{red!60} $-0.605$ & \cellcolor{red!52} $-0.528$ & \cellcolor{red!57} $-0.578$ & 
	\cellcolor{red!11} $-0.110$ & \cellcolor{red!17} $-0.174$ & \cellcolor{red!21} $-0.218$ & 
	\cellcolor{red!15} $-0.154$ & \cellcolor{red!26} $-0.265$ & \cellcolor{red!27} $-0.270$ & 
	\cellcolor{green!4} $0.048$ & \cellcolor{red!8} $-0.081$ & \cellcolor{green!0} $0.002$ \\
	\hline
	\cellcolor{gray!20} \tecnica[\ACMCMAGENTA]{MeSH} &
	\cellcolor{red!8} $-0.088$ & \cellcolor{red!5} $-0.052$ & \cellcolor{red!2} $-0.027$ & 
	\cellcolor{red!61} $-0.614$ & \cellcolor{red!48} $-0.489$ & \cellcolor{red!48} $-0.484$ & 
	\cellcolor{red!10} $-0.107$ & \cellcolor{red!20} $-0.206$ & \cellcolor{red!17} $-0.179$ & 
	\cellcolor{red!4} $-0.049$ & \cellcolor{red!22} $-0.229$ & \cellcolor{red!27} $-0.276$ & 
	\cellcolor{green!4} $0.043$ & \cellcolor{red!4} $-0.041$ & \cellcolor{red!7} $-0.078$ \\
	\hline
	\cellcolor{gray!20} \tecnica[\ACMCMAGENTA]{TF} &
	\cellcolor{green!2} $0.026$ & \cellcolor{green!0} $0.003$ & \cellcolor{green!5} $0.057$ & 
	\cellcolor{red!59} $-0.594$ & \cellcolor{red!55} $-0.555$ & \cellcolor{red!55} $-0.550$ & 
	\cellcolor{red!15} $-0.153$ & \cellcolor{red!20} $-0.208$ & \cellcolor{red!17} $-0.174$ & 
	\cellcolor{red!12} $-0.125$ & \cellcolor{red!29} $-0.291$ & \cellcolor{red!28} $-0.288$ & 
	\cellcolor{red!1} $-0.019$ & \cellcolor{red!4} $-0.046$ & \cellcolor{red!3} $-0.034$ \\
	\hline
	\cellcolor{gray!20} \tecnica[\ACMCMAGENTA]{TF-IDF} &
	\cellcolor{red!4} $-0.046$ & \cellcolor{red!2} $-0.027$ & \cellcolor{red!4} $-0.045$ & 
	\cellcolor{red!58} $-0.586$ & \cellcolor{red!54} $-0.545$ & \cellcolor{red!55} $-0.560$ & 
	\cellcolor{red!12} $-0.124$ & \cellcolor{red!28} $-0.284$ & \cellcolor{red!19} $-0.194$ & 
	\cellcolor{red!13} $-0.131$ & \cellcolor{red!22} $-0.230$ & \cellcolor{red!22} $-0.224$ & 
	\cellcolor{red!6} $-0.062$ & \cellcolor{red!5} $-0.057$ & \cellcolor{red!5} $-0.055$ \\
	\hline

        \end{tabular}
        \end{adjustbox}
        \caption{Comparison of the results of confusion tests with respect to Table \ref{tab:main-results}.}
        \label{tab:results-confusion-comparison}
        \end{table}

\section{Discussion}
\label{sec:discussion}

The general trend indicates that prompting yields modest yet consistent ROUGE improvements across all smaller models (\tecnica[\ACMCBLUE]{LED}, \tecnica[\ACMCBLUE]{LT5-Base-Local}, and \tecnica[\ACMCBLUE]{LT5-Base-ETC}) for \tecnica[\ACMCRED]{I+D}, suggesting that supplied terms offer valuable global context. However, more substantial gains are evident in the case of section-level summarization for these models. For example, \tecnica[\ACMCBLUE]{LT5-Base-ETC} demonstrates an increase of up to 0.241 in ROUGE-1 and 0.462 in ROUGE-2 when prompts are utilized for section summarization. While other techniques do not exhibit such significant enhancements, they consistently manifest improvements.

As previously stated, prompting results in significantly more significant improvements for smaller models compared to larger ones (\tecnica[\ACMCBLUE]{BigBirdPegasus} and \tecnica[\ACMCBLUE]{LT5-Large-ETC}). The substantial enhancements observed in smaller models imply that lightweight architectures, characterized by reduced representational power, exhibit a heightened dependence on supplementary external information derived from prompts that concentrate on content.

In general, no single technique consistently outperforms across all settings, implying that the optimal selection depends on specific architectures and tasks. Nevertheless, \tecnica[\ACMCMAGENTA]{KeyBERT} generally exhibits superior performance, albeit by a slight margin. This observation suggests that prompting techniques provide additional information that aids in contextualizing the summarization task. This information proves beneficial across various cases without a discernible dependence on any particular prompting technique becoming apparent.

Furthermore, the ETC attention mechanism (\tecnica[\ACMCBLUE]{LT5-Base-ETC}) appears to outperform the sliding window attention mechanism (\tecnica[\ACMCBLUE]{LT5-Base-Local}). This superiority is likely attributed to the positioning of the instruction at the beginning of the text. The sliding window attention mechanism exhibits limitations as the summary generation progresses. In contrast, utilizing global and random attention, ETC mitigates this limitation, thereby accounting for the superior results obtained. Similarly, \tecnica[\ACMCBLUE]{LED} also incorporates a global attention mechanism, underscoring the importance of adopting an attention architecture that ensures continuous access to the instruction throughout the summarization process. Notably, although \tecnica[\ACMCBLUE]{BigBirdPegasus} employs the most complex attention mechanism among the five, this complexity does not necessarily translate into enhanced performance in exploiting the studied techniques. This discrepancy may be attributed to the reduced reliance on contextual information for generating high-quality summaries, as previously discussed.

To further analyze models' reliance on informative prompts, Tables \ref{tab:results-confusion} and \ref{tab:results-confusion-comparison} present the confusion test results for section summarization, as detailed in Section \ref{sec:results:metrics}. As depicted in the tables, smaller models demonstrate significant declines in quality when exposed to shuffled prompts, while \tecnica[\ACMCBLUE]{BigBirdPegasus} and \tecnica[\ACMCBLUE]{LT5-Large-ETC} exhibit a mixture of improvements and deteriorations, showing a tendency to disregard prompts in generating predictions.

The overall trend suggests that supplying additional terms through concise prompts allows smaller models to better identify and concentrate on salient concepts to include in generated outputs. Our findings demonstrate the efficacy of prompting techniques for scientific summarizers, with implications emerging:

\begin{itemize}
    \item Prompting provides consistent gains across smaller models when summarizing sections. This confirms that prompts can focus systems on key concepts.
    \item Confusion testing reveals diminished performance when prompts are unrelated for smaller models. In this case, prompting is actively exploited rather than ignored, unlike in larger models, which benefit less from prompts, instead relying on internal learned representations.
    \item Among prompt generation techniques, \tecnica[\ACMCMAGENTA]{KeyBERT} performs marginally better. performs marginally better. Nonetheless, optimal techniques vary depending on the setting.
    \item The \Gls{ETC} attention mechanism presents advancements compared to sliding window attention. This improvement is likely attributable to the capability to integrate contextual information from the prompt through global and random attention mechanisms.
    \item For section summarization, supplying section types further improves results, confirming this extra explicit context helps.
\end{itemize}

The central implication derived from our findings is that employing decoder prompting offers a means to ameliorate the inherent limitations of smaller summarization models within suitable contexts. Rather than exclusively focusing on developing larger architectures, compact models enhanced with instructive prompts may present a practical alternative for environments with resource constraints, such as mobile devices. By addressing their inherent shortcomings through external guidance, achieving robust performance from lightweight summarizers seems attainable.

Our work introduces prompting as a general technique to meaningfully enhance small neural network summarizers. Prompting could potentially aid simple extractive systems by focusing selections and representing an alternative research direction. Our techniques provide an easily adoptable means of upgrading summarizers' capacities without requiring extensive re-engineering. Numerous promising opportunities exist for further exploration of decoder prompting.

\section{Future Work}
\label{sec:future-work}

While results indicate consistent summarization improvements from prompting techniques, especially for smaller models run on sections, there remain promising opportunities for future work. We highlight some high-level directions:

%\subsection{Additional Prompting Techniques}
\begin{itemize}
    \item Our results showed some variability in performance between different prompting techniques. Further techniques could be explored, such as RAKE \cite{rake}, TextRank \cite{textrank}, and YAKE \cite{yake}, to automatically extract salient concepts from texts to form prompts. Comparing these approaches could reveal even better-performing options.

%\subsection{Automated Prompt Generation}
    \item The approach in \cite{EntityPrompts} proposes instruction based on predicted named entities in summaries. While highly challenging for long scientific papers, future work could investigate automatically generating quality entity prompt chains. This could produce more optimized prompts than our current term extraction methods.

%\subsection{Attention to Prompt Tokens}
    \item The results indicate that sliding window attention does not perform on par with ETC for prompting. Therefore, an idea is to adapt global attention to directly focus on prompt token positions, thus avoiding the lossy aggregation into chunk representations in ETC. This adaptation could enhance prompt utilization.
\end{itemize}

In summary, there exist opportunities for future research to build upon our introduced techniques. Continued analysis of optimal methods and architectural integration merits exploration.

\section{Conclusion}
\label{sec:conclusion}

This paper introduces and assesses a collection of innovative prompting techniques aimed at enhancing scientific summarization systems by offering contextual guidance through informative key-term prompts. We propose and examine various prompting methods based on provided terms (\tecnica[\ACMCMAGENTA]{Keywords} and \tecnica[\ACMCMAGENTA]{MeSH}), salient items from \tecnica[\ACMCMAGENTA]{KeyBERT}, and extractive statistics (\tecnica[\ACMCMAGENTA]{TF} and \tecnica[\ACMCMAGENTA]{TF-IDF}). Our techniques are designed to be seamlessly integrated with any summarizer without requiring complex additional generative models or knowledge of the text to be generated, unlike in prior research.

We conducted experiments to assess the impact of our proposed prompting approaches on several state-of-the-art transformer-based scientific article summarizers, employing various model sizes and attention mechanisms. Our analysis covered both \tecnica[\ACMCRED]{I+D} and section-level summarization. Our findings indicate that focused local contexts derive the greatest benefit from global information provided through prompts. Additionally, we observed a more significant improvement in smaller models, which lack representational capacity, compared to larger models with fewer intrinsic limitations when subjected to prompting.

The findings of our study reveal consistent enhancements in standard quality metrics such as ROUGE when incorporating prompting in smaller models, particularly when executed on individual sections. These configurations yield more precise and comprehensive summaries that encompass a greater number of key concepts from gold references than unaltered models. Additionally, confusion testing provides further evidence that these models actively leverage supplied informative terms.

This work makes multiple key contributions. We introduce a new direction of decoder prompting for enhancing summarization, analyze efficacy across models and tasks, and demonstrate particular utility for improving fundamental deficiencies of smaller models in appropriate contexts. Rather than solely bigger architectures, smaller prompted models may suffice in some real-world resource-limited applications.

Our findings reveal prompting as an easily adoptable technique to upgrade base summarizer implementations designed for long-form documents. There remain open questions for future investigation, including tailoring prompts to models, developing augmented prompting mechanisms, studying sentence-level prompting, and integrating automatic evaluation of contextual relevance. There are broad opportunities for future advancements building on this work on decoder prompting to enhance the summarization of scientific articles.

\section*{Acknowledgments}

This work has received financial support from the Consellería de Educación, Universidade e Formación Profesional (accreditation 2019-2022 ED431G-2019/04), the European Regional Development Fund (ERDF), which acknowledges the CiTIUS - Centro Singular de Investigación en Tecnoloxías Intelixentes da Universidade de Santiago de Compostela as a Research Center of the Galician University System, and the Spanish Ministry of Science and Innovation (grants PDC2021-121072-C21 and PID2020-112623GB-I00). Aldan Creo is supported by the Spanish Ministerio de Universidades under the Collaboration Fellowships in University Departments scheme (BDNS-633225). Furthermore, the authors also wish to thank the supercomputing facilities provided by CESGA.

\clearpage

\printglossaries

\clearpage

\bibliographystyle{plainnat}
\bibliography{export}

\end{document}